\documentclass[conference]{IEEEtran}
\IEEEoverridecommandlockouts

\usepackage{url}
\usepackage{times}
\usepackage{amsmath}
\usepackage{epsfig}
\usepackage{multirow}
\usepackage{longtable}
\usepackage{amsfonts}
\usepackage{amssymb}%
\usepackage{color}
\usepackage{bm}
\usepackage{epstopdf}
\usepackage{subfigure}
\usepackage{cite}
\usepackage{graphicx}
\usepackage{amsmath}
\usepackage{times}
\usepackage{latexsym}
\usepackage[center]{caption2}
\usepackage{stfloats}
\usepackage{cases}
\usepackage{array}
\usepackage{setspace}
\usepackage{fancyhdr}
\usepackage{balance}
\usepackage{algorithm} 
\usepackage{algpseudocode} 

\usepackage[table]{xcolor}
\usepackage{makecell}

        {\par\noindent{\bf Proof.}}%
        {{\hfill{\large$\Box$}}\smallskip}

\begin{document}

\title{SREC: Proactive Self-Remedy of Energy-Constrained UAV-Based Networks via Deep Reinforcement Learning
\thanks{This work was supported by Natural Science Foundation (NSF) ECCS under Grant 1554576.}
}


\author{\IEEEauthorblockN{
Ran Zhang\IEEEauthorrefmark{1},
Miao Wang\IEEEauthorrefmark{1},
and Lin X. Cai\IEEEauthorrefmark{2}}\\
\IEEEauthorblockA{\IEEEauthorrefmark{1}
Department of Electrical and Computer Engineering,
Miami University, Oxford, USA}
\IEEEauthorblockA{\IEEEauthorrefmark{2}
Department of Electrical and Computer Engineering,
Illinois Institute of Technology, Chicago, USA}
\IEEEauthorblockA{Email: \IEEEauthorrefmark{1}\{$zhangr43,wangm64$\}@miamioh.edu,  \IEEEauthorrefmark{2}$lincai$@iit.edu}
}

\maketitle

\begin{abstract}
Energy-aware control for multiple unmanned aerial vehicles (UAVs) is one of the major research interests in UAV based networking. Yet few existing works have focused on how the network should react around the timing when the UAV lineup is changed. In this work, we study proactive self-remedy of energy-constrained UAV networks when one or more UAVs are short of energy and about to quit for charging. We target at an  energy-aware optimal UAV control policy which proactively relocates the UAVs when any UAV is about to quit the network, rather than passively dispatches the remaining UAVs after the quit. Specifically, a deep reinforcement learning (DRL)-based self remedy approach, named SREC-DRL, is proposed to maximize the accumulated user satisfaction scores for a certain period within which at least one UAV will quit the network. To handle the continuous state and action space in the problem, the state-of-the-art algorithm of the actor-critic DRL, i.e., deep deterministic policy gradient (DDPG), is applied with better convergence stability. Numerical results demonstrate that compared with the passive reaction method, the proposed SREC-DRL approach shows a $12.12\%$ gain in accumulative user satisfaction score during the remedy period.
\end{abstract}

\section{Introduction}\label{sec.Intro}
Unmanned aerial vehicles (UAVs) have been demonstrating dazzling potentials in future wireless communications due to flexible 3-D mobility and higher chance of Line-of-Sight (LoS) links to ground users compared to the terrestrial base stations (BSs)\cite{zeng2016wireless}. UAVs equipped with wireless transceivers have found applications in a multitude of fields, such as emergency rescue, UAV-based heterogeneous networks (HetNets), rural Internet service support, traffic monitoring and control, cached content delivery, etc.. In various applications, UAVs serve as either relays to collect or disseminate data, or additional access points to improve the communication performance.

While having a bright future, UAV-based communications still face a number of research challenges. Many related works have been conducted in various aspects, such as resource allocation, deployment, and trajectory design\cite{hu2018uav,zhao2019uav,wu2018joint}. 
Compared with those studied in terrestrial networks, UAV-based problems have extra dimensions due to UAV mobility. Particularly, for UAV trajectory design, the target performance metric usually considers UAV and/or user dynamics for a range of time slots, thus being sophisticated especially when multiple UAVs need coordinating. The conventional optimization or rule-based approaches have been increasingly labored when dealing with such problems. Thanks to the recent advance in machine learning, reinforcement learning (RL) and deep RL (DRL) techniques\cite{luong2019applications} are becoming promising solutions to UAV-based problems\cite{klaine2018distributed,singh2018distributed,liu2019optimized}. Bearing the nature of maximizing accumulative future return, (D)RL is strongly capable of handling complex state and action space in time-varying environment with limited to even zero domain knowledge. For instance, Klaine \emph{et. al.} \cite{klaine2018distributed} proposed a distributed Q-learning approach to find best UAV positions that maximize the total amount of covered users with time-varying user distributions. Singh \emph{et. al.} \cite{singh2018distributed} exploited Deep Q-learning (DQL) to optimize the inter-cell interference coordination in UAV-based HetNets. Liu \emph{et. al.} \cite{liu2019optimized} employed double Q-learning to design optimal trajectories for multiple UAVs which maximize the number of satisfied users with time-constrained service requirements.

In addition to the space mobility, energy management is another major research interest for UAV-based networks as UAVs are usually powered by battery. Quite a few research works have considered energy-related UAV control.
For instance, Sun \emph{et. al.} \cite{sun2019optimal} designed optimal 3D trajectories for a single solar-powered UAV and identified the tradeoff between communication performance and solar energy harvesting. Liu \emph{et. al.} \cite{liu2018energy} exploited the DRL approach to devise an optimal UAV control policy that jointly maximizes the energy efficiency, fairness and coverage performance of UAV networks.
However, there are few works in the existing literature that focus on how the network should react around the timing when one or more UAVs are short of energy and demand to leave for charging. By ``react", only passive respond after any UAV quits the network is not enough, the network is anticipated to identify the upcoming change in UAV lineup and proactively take effective actions in advance to minimize performance loss during the transition to another steady state. Such procedure is referred to as proactive self-remedy in this paper.

Therefore, in this work, we study proactive self-remedy of energy-constrained UAV-based networks when one or more UAVs are running out of battery and about to quit for charging. We target at an energy-aware optimal UAV control policy which proactively relocates the UAVs when any UAV is about to quit the network, rather than passively dispatches the remaining UAVs after the quit. Specifically, the contributions of the paper are summarized as follows.
\begin{itemize}
    \item A DRL-based approach for self-remedy of energy constrained UAV networks, named as SREC-DRL, is proposed. The approach is designed to maximize the accumulated user satisfaction scores for a certain period within which at least one UAV will quit the network. DRL techniques are employed due to its policy-oriented nature of maximizing the accumulated future return.
    \item The SREC-DRL approach identifies the UAV positions and battery status as states, moving directions and distances as actions, and user satisfaction score as instantaneous reward. As energy-aware UAV control is a continuous problem, the state-of-the-art actor-critic method, Deep Deterministic Policy Gradient (DDPG)\cite{lillicrap2015continuous}, is applied as an implementation of DRL to handle the continuous state and action space.
    \item Numerical results are presented to demonstrate the efficacy of the proposed approach. Compared with the passive reaction method, SREC-DRL shows a $12.12\%$ gain during the remedy period.
\end{itemize}


The reminder of the paper is organized as follows. Section \ref{sec.SystemModel} describes the system model. Section \ref{sec.PK} introduces preliminary knowledge on DRL and the applied DDPG algorithm. Section \ref{sec.Algorithm} elaborates the detailed design of the proposed SREC-DRL approach. Numerical results are presented in Section \ref{sec.Simulation}. Finally, Section \ref{sec.Conclusion} concludes the paper.

\section{System Model}\label{sec.SystemModel}
In this section, the system model is depicted from four aspects: network setup, spectrum access, energy-related setup, and system objectives.
\subsection{Network Setup}\label{subsec.nt}
\begin{figure}[!ht]
	\centering
	\includegraphics[width=3.4in]{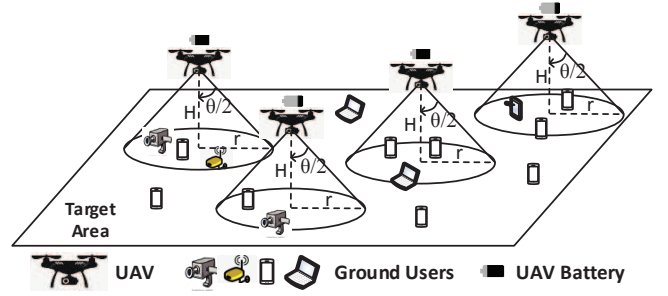}
	\caption{UAV coverage range as a disk area.} \label{fig.SystemModel}
\end{figure}

We consider a target area $\mathbf{A}$ with a set $\mathbf{S}_{ur}$ of $N_u$ ground users, as shown in Fig. \ref{fig.SystemModel}. The target area is (but not limited to) an $L$-by$L$ square. A percentage $p$ of the users are randomly distributed around certain scattered hot spots while the rest are uniformly distributed throughout $\mathbf{A}$. A group $\mathbf{S}_{UAV}$ of $N_{UAV}$ UAVs fly horizontally within $\mathbf{A}$ at a fixed altitude $H$ to provide QoS guaranteed communications to the ground users. The antennas of each UAV are strongly directional, such that most of the energy is focused within an aperture angle of $\theta$ right below the UAV. Thus the ground coverage $\mathbf{R}$ of a UAV is considered to be a disk area with radius $r=H\tan(\frac{\theta}{2})$ as shown in Fig. \ref{fig.SystemModel}. Users will not receive interference from one UAV if they are outside its coverage disk.

\subsection{Spectrum Access}\label{subsec.sa}
All the UAVs are considered to have connections (e.g., satellite links) to external networks (e.g., Internet). These links are set to occupy disjoint spectrum from that of UAV-user links, thus having no mutual interference. The path loss from UAV $i$ to ground user $u$ follows a commonly adopted model proposed by Al-Hourani \emph{et al.}\cite{al2014modeling} as given below,
\begin{equation}\label{eq.PL}
PL_{iu} = 20\log_{10}{(\frac{4\pi f_cd_{iu}}{c})}+\eta \;\;\; \text{(dB)},
\end{equation}
where $f_c$ denotes the center frequency of the occupied spectrum, $d_{iu}$ denotes the 3-D distance from UAV $i$ to user $u$, $c$ denotes the light speed, and $\eta$ denotes additional loss which has different values for LoS or non-LoS links. With the considered path loss, the siginal-to-interference-and-noise ratio (SINR) from UAV $i$ to user $u$ is obtained as
\begin{equation}\label{eq.sinr}
\begin{array}{l}
    SINR_{iu} = \frac{P_tG_{iu}}{n_0+\sum_{j\in {\mathbf{S}_u}'\backslash\{i\}}P_tG_{ju}}\\
    \text{where }G_{iu}=10^{-PL_{iu}/20}.
\end{array}
\end{equation}
In Eq. \eqref{eq.sinr}, $P_t$ and $n_0$ is the power spectrum density (psd) for UAV transmission and noise, respectively; ${\mathbf{S}_u}'$ denotes the set of UAVs having user $u$ in their coverage.

Each user is considered to have a minimum throughput requirement $r_{u}$. A user can access a UAV only when its QoS requirement can be satisfied, i.e.,
\begin{equation}\label{eq.access}
W_{iu}\log_2{(1+SINR_{ui})}\ge r_u,
\end{equation}
where $W_{iu}$ denotes the bandwidth assigned by UAV $i$ to the user. Each user is allocated to the UAV that provides the best SINR with enough available bandwidth, according to Eq. \eqref{eq.access}.

\subsection{Energy-Related Setup}\label{subsec.energy}
Each UAV $i$ is battery-powered with initial energy $E^i_{0}$. The timing is divided into a sequence of time slots, each with duration $T$. In each time slot $t$, each UAV spends at most $T_1<T$ to move a distance of $d^i_t\in [0,d_{max}]$ at a constant speed $v$ in the direction of $\alpha^i_t\in[0,2\pi)$, and hovers in the new position for the rest of time to interact with the environment. The induced power of level flight is given as follows according to \cite{seddon2011basic} Eq. (7.10),
\begin{equation}\label{eq.level}
P_{level}=\frac{W}{\sqrt{2}\rho A}\frac{1}{\sqrt{v^2+\sqrt{v^4+4V_h^4}}},
\end{equation}
where $V_h=\sqrt{\frac{W}{2\rho A}}$, $W$ is the weight of UAV in Newton ($N$), $\rho$ is the air density, and $A$ is the total area of UAV rotor disks. It can be inferred that the power of level flight is interestingly less than that of hovering due to the speed. Therefore, the energy consumption $EC^i_t$ for UAV $i$ in time slot $t$ is given as
\begin{equation}\label{eq.energy}
EC^i_t = E_{FLT}(v,d^i_t,T) + E_{TX}(W^i_t,P_t,T) + E_{OP}(T).
\end{equation}
The energy consumption of a UAV is divided into three parts: one part due to flight as a function of level speed $v$, flight distance $d^i_t$ and slot duration $T$, one part due to signal transmission as a function of total occupied bandwidth $W^i_{t}$, psd $P_t$ and $T$, and one part due to operational cost assumed to be proportional to $T$. Denote the residual battery energy of UAV $i$ at the end of time slot $t$ as $E^i_t$. When $E^i_t$ is below a threshold $E_{Thre}$, UAV $i$ will quit the network for charging.

\subsection{System Objectives}\label{subsec.obj}
The system aims to find a control policy that maximizes the accumulative user satisfaction scores within a period of $N_T$ time slots, given the user locations and energy status of each UAV. Particularly, when one UAV is running out of battery and about to quit, the system is expected to take actions in advance before the quit to reduce service holes as much as possible, rather than start dispatching after the UAV quits.

The user satisfaction score $SC_{t}$ at time slot $t$ is defined as:
\begin{equation}
    SC_{t}:=\left(\sum_{u\in\mathbf{S}_{ur}}{X^u_t}\right)^\beta,
\end{equation}
where $X^u_t\in\{0,1\}$ takes $1$ when user $u$ is served with satisfied QoS requirement and $0$ when not, and $\beta>0$ is a factor representing how much we care about the user satisfaction relative to the number of users that get served in time slot $t$.
\section{Preliminaries}\label{sec.PK}

In a general reinforcement learning (RL) context, an RL agent interacts with a system environment and gets a reward $r_t$ for taking an action $A_t$ in a system state $S_t$ at epoch $t$. The agent targets at a policy $\pi$ which determines the best action $A$ for different system states $S$ to maximize the future cumulative long-term rewards $R$ defined as
$R = \sum_{t=0}^{\infty}{\gamma^t r_{t+1}},\;\gamma\in[0,1].$
Q-learning (QL)\cite{han2020} is the most commonly used value-based method to get the optimal $\pi$. Instead of optimizing $\pi$ directly, QL embarks on evaluating the value function $Q(S,A)$ of taking action $A$ at state $S$. Then the optimal policy is $\pi^*(S)=\arg\max\limits_AQ(S,A)$.
From the Bellman Optimality Equation\cite{he2018reinforcement}, $Q(S,A)$ can be updated with guaranteed convergence as follows,
\begin{equation}
\begin{array}{l}
    Q_{t+1}(S_t,A_t) = Q_t(S_t,A_t)\\ +\alpha\left[r_{t+1}+\gamma\max\limits_{A}{Q_t(S_{t+1},A)}-Q_t(S_t,A_t)\right],
\end{array}
\end{equation}
where $t$ is the epoch and $\alpha$ is the tunable learning rate. One of the major drawbacks of QL is that the agent needs to maintain a $Q$-matrix for each possible state and action, which will suffer from ``curse of dimensionality" when the state space is large or even infinite. This is often the case in the field of communications and networking. In light of this, DRL is proposed exploiting one or multiple deep neural networks (DNNs) as approximator(s) of the $Q(\cdot)$ function and/or the action decider. As a representative, deep QL (DQL) employs a DNN, referred to as deep Q-network (DQN), to approximate the $Q(\cdot)$ function. Compared to the $Q$-matrix, DQN makes the input dimension equal to the cardinality of the state space, thus solving the memory anxiety of the standard Q-learning. The DQN is trained by minimizing the loss function below\cite{hester2018deep}:
\begin{equation}\label{eq.loss}
    \mathcal{L}(\theta_Q)=\mathbb{E}[y_t-Q(S_t,A_t|\theta_Q)]^2,
\end{equation}
where $\theta_Q$ is the weight set of DQN, $y_t$ is the target value given by
\begin{equation}\label{eq.target1}
    y_t = \left\{
    \begin{array}{l}
         r_{t+1},\text{ if }S_t\text{ is terminal state;}\\
         r_{t+1}+\gamma \max\limits_{A_{t+1}}Q(S_{t+1},A_{t+1}|\theta_Q),\text{ otherwise.}
    \end{array}\right.
\end{equation}

DQL only applies to problems with low-dimensional discrete action space. For a variety of problems in communication field such as power control and UAV movement control (considered in this paper), the action space is continuous. Discretizing the action space is feasible, but resulting in extremely long training period and/or considerable accuracy loss. Actor-critic RL (AC-RL)\cite{grondman2012survey} well solves the problem by employing DNNs to approximate both the $Q(\cdot)$ function (critic) and the action decider (actor). Instead of DQN, the actor network is used to determine the action and updates itself in a policy gradient manner.

Among all the AC-RL algorithms, DDPG is considered one of the best in solving the convergence instability problems\cite{lillicrap2015continuous}.
Specifically, DDPG adopts target networks for both the critic network ($Q(S,A|\theta_Q)$) and actor network ($\mu(S|\theta_\mu)$). The target networks, denoted as $Q'(S,A|\theta_{Q'})$ and $\mu'(S|\theta_{\mu'})$, have the same hyper-parameter setting and initialization as $Q(S,A|\theta_Q)$ and $\mu(S|\theta_\mu)$, respectively, but are updated slowly in each epoch as follows:
\begin{equation}\label{eq.target}
\begin{array}{l}
    \theta_{Q'} = \tau\theta_Q+(1-\tau)\theta_{Q'}\\
    \theta_{\mu'} = \tau\theta_\mu+(1-\tau)\theta_{\mu'}
\end{array}
\end{equation}
where $\tau<<1$. Such slow update prevents the target value $y_t$ from being affected too much by a bad deviation in $\theta_{Q}$ or $\theta_{\mu}$, thus significantly improving the convergence stability. With DDPG, the target $y_t$ in Eq. \eqref{eq.target1} is estimated by
\begin{equation}\label{eq.losstarget}
    y_t = \left\{
    \begin{array}{l}
         r^{t+1},\text{ if }S_t\text{ is terminal state;}\\
         r_{t+1}+\gamma Q'(S_{t+1},\mu'(S_{t+1}|\theta_{\mu'})|\theta_{Q'}),\text{ otherwise.}
    \end{array}\right.
\end{equation}
The actor network is updated as follows,
\begin{equation}\label{eq.actor}
    \begin{array}{l}
        \nabla_{\theta_\mu}J \approx \mathbb{E}[G_{a}G_{\mu}],\\
        \text{where }G_{a}=\nabla_{\mu(S|\theta_\mu)}Q(S,A|\theta_Q),\\
        \;\;\;\;\;\;\;\;\;\;G_{\mu}=\nabla_{\theta_\mu}\mu(S|\theta_\mu).
    \end{array}
\end{equation}
\section{Design of SREC-DRL Approach}\label{sec.Algorithm}
In this section, detailed design of the proposed SREC-DRL approach is elaborated. A well-trained centralized SREC-DRL agent periodically collects the states of the UAV-based network in each time slot (i.e., epoch), finds the best action to take, and directs the UAVs to move via the backhaul links of UAV. To train the agent, the states, actions, state transitions, reward functions, and tune-ups are explained as follows.

\subsection{State Definition}
The system has three groups of states: UAV positions, residual battery energy of each UAV, and whether each UAV has quit the network.
\begin{itemize}
    \item The positions of UAVs directly affect the number of served users in each epoch, thus largely affecting the maximization objectives. As the UAVs fly horizontally, we only care about the 2-D coordinates $(x^i_t,y^i_t)$, $\forall i\in\mathbf{S}_{UAV}$ at epoch $t\le N_T$. The movements of UAVs are bounded within the target area $\mathbf{A}$, so $x^i_t,\;y^i_t\in[0,L]$.
    \item The residual battery energy of UAVs is another key factor. It is expected to have little impact on the UAV movement when the energy of all UAVs are abundant. But when the battery of any UAV is close to $E_{Thre}$, this factor is expected to have significant influence on the movement of UAVs. The SREC-DRL agent will learn from the experiences the critical timing for $\{E^i_t\}$ to take part in the game. Moreover, $E^i_t$ is bounded in $[E_{Thre},E^i_0]$.
    \item Whether one UAV quits the network is also involved as part of the environment states. Let $q^i_t$ be the indicator of whether UAV $i$ quits the network at epoch $t$, with value 1 indicating ``not quit", and 0 indicating ``quit". Although the residual battery energy \emph{implicitly} indicates $q^i_t$, lack of this information as \emph{explicit} states will significantly increase the non-linearity of the learning task, leading to convergence at local optimum.
\end{itemize}
Collectively, the formal state vector of SREC-DRL is defined as $S_t=[x^1_t,\cdots,x^{N_{UAV}}_t,y^1_t,\cdots,y^{N_{UAV}}_t,E^1_t,\cdots,E^{N_{UAV}}_t,$ $q^1_t,\cdots,q^{N_{UAV}}_t]$, with cardinality of $4N_{UAV}$.

\subsection{Action Definition}\label{subsec.action}
As the training takes place in a centralized agent, the action $A_t$ of SREC-DRL in epoch $t$ corresponds to collective actions from all the UAVs. The action of each UAV consists of two factors: moving direction $\alpha^i_t\in[0,2\pi)$ and moving distance $d^i_t\in[0,d_{max}]$. In other words, each UAV could keep still in its current position or move in any direction with a maximum distance $d_{max}$. Therefore, the formal action vector of SREC-DRL is defined as $A_t=[\alpha^1_t,\cdots,\alpha^{N_{UAV}}_t,d^1_t,\cdots,d^{N_{UAV}}_t]$, with cardinality of $2N_{UAV}$.

\subsection{State Transition}

The terminal state is largely coupled with the definition of the instantaneous reward. In SREC-DRL, one episode ends in two situations: $i$) whenever any UAV moves out of boundaries, i.e., $x^i_t$ (or $y^i_t$) $<0$ or $x^i_t$ (or $y^i_t$) $>L$; $ii$) when $t$ reaches the period end $N_T$. When reaching the terminal state, the agent ends the current episode and starts a new one. The objective is to maximize the accumulated reward of one episode.

The state space and action space will be different after one UAV quits the network. Suppose UAV $i$ quits the network at epoch $t_q$. Then $x^i_t$, $y^i_t$, $E^i_t$ and $q^i_t$ keep unchanged for any $t>t_q$. For actions, no matter what values of $\alpha^i_t$ and $d^i_t$ are suggested by the actor network, $d^i_t$ is always treated as 0; UAV $i$ is always excluded when calculating the reward.

\subsection{Reward Function Design}
Denote the reward for epoch $t$ as $r_t$. The epoch reward is designed as a function of user satisfaction score $SC_t$:
\begin{equation}\label{eq.reward}
    r_t = \left(\frac{\sum_{u\in\mathbf{S}_{ur}}{X^u_t}}{N_u}\right)^\beta=\frac{EC_{t}}{(N_u)^\beta}.
\end{equation}
where the numerator is the user satisfaction score in epoch $t$. We divide $EC_t$ by $(N_u)^\alpha$ because empirically speaking, it is preferable to keep the absolute value of the instantaneous reward within $1$ for better convergence performance. Moreover, when $\beta>1$, the reward difference for different ($\sum_{u\in\mathbf{S}_{ur}}{X^u_t}$) values is amplified. This is to promote the agent to take advanced actions when one UAV is about to quit the network. But $\beta$ cannot be too large as it may end up with lower converged return in practical simulations.
\begin{algorithm}[!ht]
\caption{SREC-DRL Approach}
\label{Alg:algorithm}
\begin{algorithmic}[1]
\State{Randomly initialize critic network $Q(S,A|\theta_Q)$ and actor network $\mu(S|\theta_\mu)$;}
\State{Initialize the target networks $Q'(S,A|\theta_{Q'})$ and $\mu'(S|\theta_{\mu'})$ with the same weights: $\theta_{Q'} := \theta_{Q}, \theta_{\mu'}:=\theta_{\mu}$;}
\For{episode := $1,\cdots,N$}
    \State{Obtain the initial state $S_1$, IsTerminal := False;}
    \For{epoch $t$ := $1,\cdots,N_T$}
        \State{$A_t=\mu(S|\theta_\mu)+\mathcal{N}$, where $\mathcal{N}$ is stochastic noise $\text{ }\text{ }\text{ }\text{ }\text{ }\text{ }\text{ }\text{ }\text{ }$with zero mean and decaying variance over $t$;}
        \State{Execute $A_t$ and observe next state $S_{t+1}$;}
        \For{UAV $i$ := $1,\cdots,N_{UAV}$}
            \If{UAV $i$ has quit the network}
                \State{$S^i_{t+1}:=S^i_t$, where $S^i_t$=$\{x^i_t,y^i_t,E^i_t,q^i_t\}$;}
                \State{Exclude UAV $i$ when calculating $r_{t+1};$}
            \EndIf
            \If{UAV $i$ goes out of boundaries}
                \State{Cancel the movement of UAV $i$;}
                \State{IsTerminal := True;}
            \EndIf
        \EndFor
        \State{Calculate $r_{t+1}$;}
        \State{Store the experience $(S_t, A_t, r_{t+1},S_{t+1})$ into $\text{ }\text{ }\text{ }\text{ }\text{ }\text{ }\text{ }\text{ }\text{ }$experience replay buffer $B$;}
        \State{Obtain a random minibatch of samples from $B$;}
        \State{Update $\theta_Q$ according to Eq. \eqref{eq.loss}\eqref{eq.losstarget};}
        \State{Update $\theta_{\mu}$ according to Eq. \eqref{eq.actor};}
        \State{Update $\theta_{Q'}$ and $\theta_{\mu'}$ according to \eqref{eq.target};}
        \State{Break if IsTerminal==True;}
    \EndFor
\EndFor
\end{algorithmic}
\end{algorithm}

An alternative design of reward function is to assign negative rewards as punishment when any UAV moves out of boundaries. The reward function will be something like:
\begin{equation}
    r_t = \left\{
    \begin{array}{l}
    \left(\sum_{u\in\mathbf{S}_{ur}}{X^u_t}/{N_u}\right)^\beta,\;\;\text{if inside boundaries}\\
    p,\;\;\text{otherwise}
    \end{array}
    \right.
\end{equation}
where $p$ can be a negative constant or negatively proportional to the number UAVs outside the boundaries. Accordingly, when one UAV moves out of boundaries, the movement will be cancelled, and a punishment will be issued. As a result, each episode will all have $N_T$ epochs. While being feasible, this design may be more difficult towards convergence. This is because empirically a good reward design needs to make the negative rewards ''combat" the positive ones closely during the training for better convergence performance. However, the percentages of positive and negative rewards keep changing during the training, thus very likely taking more iterations to tune the relative values between positive and negative rewards.

\subsection{Agent and Training Tune-Ups}
The critic and actor DNNs should be just large enough to well learn the input-output mappings and prevent overfitting. Both the critic and actor DNNs exploit 2 fully connected hidden layers, each having 400 and 300 neurons, respectively. To bound the actions as defined in Subsection \ref{subsec.action}, \emph{tanh} and \emph{scaling} layers are used in the actor DNN. For both networks, ReLU function and $L_2$ regularization is used for activation and overfitting prevention, respectively.  The learning rates for updating both DNNs is $10^{-4}$. A larger learning rate in our case tends to result in convergence instability or local optimum. The mini-batch size is $512$ and input normalization is adopted.

As for RL training, both target networks are updated slowly with $\tau=0.001$. DDPG exploits an exploration algorithm where the output of the actor network is added with a random noise of zero mean and decaying variance. In our implementation, the initial variance is 0.6 and decays at a rate of 0.9995. The discount factor $\gamma$ is set to 0.9. A higher $\gamma$ forces the agent to consider the future rewards more, thus making convergence harder. Experience replay is adopted with enough buffer size for all the experiences. In our implementation, insufficient buffer size will make the agent lose failure experiences at early stage, subsequently causing notable instability during training. The SREC-DRL approach is presented in Algorithm \ref{Alg:algorithm}.

\section{Numerical Results}\label{sec.Simulation}

\subsection{Simulation Setup}
The target area is square with $10\times10$ square units. Each unit is 100 meters. The training is conducted using Reinforcement Learning Toolbox of Matlab 2020a on a Windows 10 server with Intel Core i7-7700 CPU @ 3.60GHz and 16GB RAM. The training has maximum 10000 episodes, each having up to 100 epochs. The trained agent is tested for a period of $N_T=100$ epochs. Moreover, the communication-related power of UAVs is considered negligible compared to the flight/hovering power. The main parameters are summarized in Table \ref{Table:notation} below\footnote{Note that $unit\cdot s$ in the table indicates that the value is a product of power (1 power unit = 9.428W according to \eqref{eq.level}) and time (unit is second).}.

\begin{table}[!ht]
\footnotesize
\centering
\renewcommand{\arraystretch}{1}

\begin{tabular}{!{\vrule width0.8pt}l|l!{\vrule width0.8pt}}\Xhline{0.8bp}
\multicolumn{1}{!{\vrule width0.8pt}c|}{\gape{\bfseries Parameters}} & \multicolumn{1}{c!{\vrule width0.8pt}}{\gape{\bfseries Values}} \\ 



  \hline
  \rowcolor[gray]{0.9}
  UAV level speed $v$ & 40$km/h$\\
  UAV weight $W$, air density & 4$kg\times$9.8$m/s^2$, 1.225$kg/m^3$\\
  \rowcolor[gray]{0.9}
  Total area of rotor disks $A$ & 0.18$m^2$\\
  UAV height $H$, apenture angle $\theta$  & 3 units, $60^o$ \\
  \rowcolor[gray]{0.9}
  Max. distance per epoch $d_{max}$ & 1 unit \\
  Spectrum center frequency $f_c$  & 2GHz\\
  \rowcolor[gray]{0.9}
  Spectrum access technology & LTE with resource blocks (RBs)\\
  Spectrum and RB bandwidth & 4.5MHz and 180kHz\\
\rowcolor[gray]{0.9}
  psd of transmission and noise& -49.5dBm, -174dBm \\
  Required user throughput $r_u$ & 250kbps\\
  \rowcolor[gray]{0.9}
  LOS path loss parameter $\eta$&1dB \\
  Time duration per epoch $T$&10s\\
  \rowcolor[gray]{0.9}
  Max. UAV moving (communication) & 9s (1s)\\
  \rowcolor[gray]{0.9}
  time per epoch $T_1$ ($T-T_1$)&~~ \\
  Factor of user satisfaction score $\beta$&2 \\
  \rowcolor[gray]{0.9}
  Energy threshold to quit $E_{Thre}$ & 150 unit$\cdot$s\\

  \hline
  \end{tabular}
\caption{Summary of Main Parameters}\label{Table:notation}
\end{table}

\subsection{Simulation Results}\label{subsec.simuresults}

\begin{figure}[!ht]
	\centering
	\includegraphics[width=3.2in]{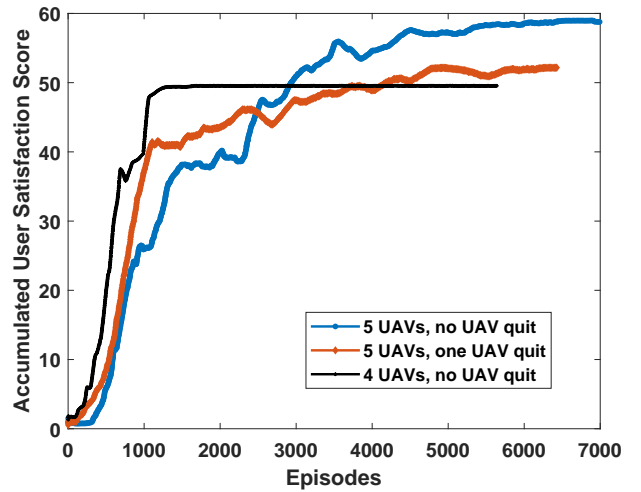}
	\caption{Episode-wise accumulated user satisfaction scores.} \label{fig.Converge}
\end{figure}

The episode-wise accumulated user satisfaction scores during training are first presented in Fig. \ref{fig.Converge}. Three configurations are considered: 5 UAVs all with sufficient battery energy, 5 UAVs of which one UAV quits the network, 4 UAVs all with sufficient battery energy. It is shown that all the three configurations converge eventually. Configurations with 5 UAVs converge much more slowly than the one with 4 UAVs. Note that what is in the y-axis is not the episode reward, but the average reward over the latest 200 episodes. The reason is that instead of converging to a single constant value, each curve eventually converges to a narrow range with $95\%$ credit interval around 3.

\begin{figure}[!ht]
	\centering
	\includegraphics[width=3.1in]{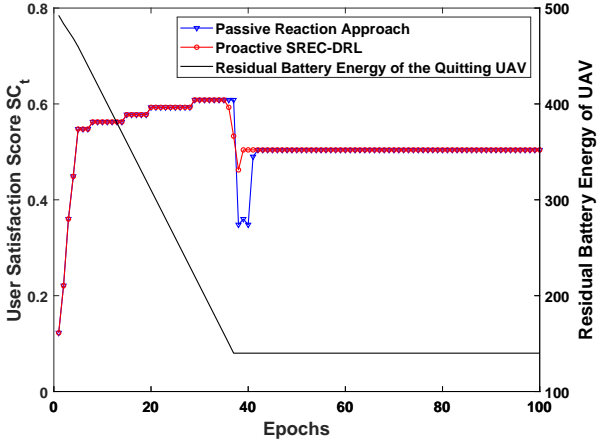}
	\caption{User Satisfaction Score Comparison and Residual Battery Energy} \label{fig.Episode}
\end{figure}

The optimal epoch-wise user satisfaction scores under both the proposed proactive approach and the passive reaction approach are presented in Fig. \ref{fig.Episode}. The residual battery energy of the only UAV that quits the network during the testing period is also provided. It can be observed that the battery energy of the considered UAV decreases slower in the first several epochs compared to the after. At beginning, UAVs are far from the optimal positions, and thus need to fly for longer time; whereas when getting close to the optimal positions, UAVs just make marginal adjustments and hover for more time. As level flight consumes less power than hovering, less energy is consumed in the beginning. In addition, the considered UAV quits the network around epoch 40 when its battery energy drops below $E_{Thre}=150$ unit$\cdot$s. More importantly, the epoch-wise user satisfaction scores under both approaches first increase to the maximum $0.6084$, then have a dramatic drop around the time of UAV quit, and finally goes up to $0.5041$ after the remaining UAVs reach the new optimal positions.

\begin{figure}[!ht]
	\centering
	\includegraphics[width=2.9in]{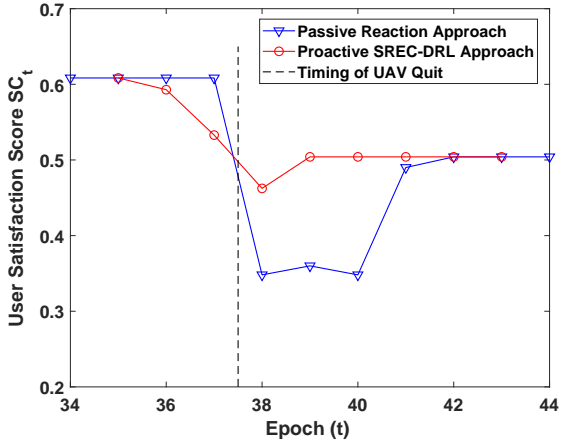}
	\caption{User Satisfaction Score Comparison in Self-Remedy Period.} \label{fig.Enlarge}
\end{figure}

A zoomed-in view of the user satisfaction scores around the timing of UAV quit is shown in Fig. \ref{fig.Enlarge}. It can be seen that under the passive reaction approach, UAVs remain still at the optimal positions until one UAV quits, and then passively start relocating. But under SREC-DRL approach, UAVs start relocating 2 epochs ahead of the UAV quit, so that the user satisfaction score does not drop too much when the UAV quits. As a result, SREC-DRL approach gains $12.12\%$ higher accumulative user satisfaction score over the passive reaction approach during the remedy period [36,41].
\section{Conclusions}\label{sec.Conclusion}
In this paper, self-remedy of the energy constrained UAV-based networks has been studied. DDPG algorithm has been applied to handle the continuous action space while maximizing the accumulative user satisfaction scores. Specifically, a DDPG agent has been successfully trained to proactively relocate the UAVs when any UAV is about to quit the network due to battery depletion, rather than passively start relocation after one UAV quits. Simulation results have demonstrated the efficacy of the proposed SREC-DRL approach by having a $12.12\%$ gain over the passive method.

\bibliographystyle{IEEEtran}
\bibliography{main}

\end{document}